\documentclass[11pt]{article} 

\usepackage[utf8]{inputenc} 

\usepackage{geometry} 
\usepackage{graphicx} 
\usepackage{tikz}
\usepackage{amsmath}
\usepackage{amssymb}
\usepackage{amsthm}
\usepackage{natbib}
\usepackage{appendix}
\usepackage[ruled,vlined]{algorithm2e}
\usepackage{framed}

\usepackage{booktabs} 
\usepackage{array} 
\usepackage{paralist} 
\usepackage{verbatim} 
\usepackage{subfig} 

\usepackage{fancyhdr} 
\usepackage{sectsty}
\allsectionsfont{\sffamily\mdseries\upshape} 

\usepackage[nottoc,notlof,notlot]{tocbibind} 
\usepackage[titles,subfigure]{tocloft} 

\newcommand{\bs}[1]{\boldsymbol#1}

\newcommand{\asg}[2]{A^*_{\gamma}\left(#1, #2 \right)}
\newcommand{\asgg}[1]{A^*_{\gamma}\left(#1 \right)}

\newtheorem{theorem}{Theorem}
\newtheorem{proposition}{Proposition}
\theoremstyle{definition}
\newtheorem{definition}{Definition}
\newtheorem{example}{Example}

\title{Approximation to Object Conditional Validity\\with Inductive Conformal Predictors}
\author{Anthony Bellotti\\
School of Computer Science, University of Nottingham Ningbo China\\
Correspondence: \texttt{Anthony-Graham.Bellotti@nottingham.edu.cn} }
\date{} 

\begin{document}
\maketitle

\abstract{
Conformal predictors are machine learning algorithms that output prediction sets that have a guarantee of marginal validity for finite samples with minimal distributional assumptions. This is a property that makes conformal predictors useful for machine learning tasks where we require reliable predictions.
It would also be desirable to achieve conditional validity in the same setting, in the sense that validity of the prediction intervals remains true regardless of conditioning on any particular property of the object of the prediction.
Unfortunately, it has been shown that such conditional validity is impossible to guarantee for non-trivial prediction problems for finite samples.
In this article, instead of trying to achieve a strong conditional validity result, the weaker goal of achieving an \emph{approximation} to conditional validity is considered. 
A new algorithm is introduced to do this by iteratively adjusting a conformity measure to deviations from object conditional validity measured in the training data.
Along with some theoretical results, experimental results are provided for three data sets that demonstrate (1) in real world machine learning tasks, lack of conditional validity is a measurable problem and (2) that the proposed algorithm is effective at alleviating this problem.
}

\section{Introduction}

In some predictive analytic tasks it is useful to provide a prediction set of possible outcomes rather than a single point prediction. For example, a doctor diagnosing a patient's symptoms may find it more valuable if an automated decision support system suggested two or more possible causes instead of just one, if there is some ambiguity in diagnosis \citep{Nouretdinov2014}. Or, given the uncertainty in property sale prices, it may be better if an automated valuation model provides a range of possible predicted sale prices, rather than a single point estimate. Indeed, for automated valuation, the delivery of a property price range is the norm; see e.g. \citep{Bellotti2017} or \texttt{www.zoopla.co.uk}. 

The conformal predictor (CP) is a machine learning algorithm that is able to generate prediction sets with a guarantee that its prediction sets are correct, i.e. the true label 
\footnote{We use the convention that each entity for which a prediction is to be made is called an \emph{example}, with the \emph{label} being the outcome we wish to predict and \emph{object} the collection of features from which we may make a prediction, following \cite{Vovk2014}.}
will be a member of the prediction set, at a preset confidence level \citep{Vovk2005AlgorithmicWorld}. We therefore say that the CP is a \emph{valid} predictor. Other statistical methods based on frequentist and Bayesian approaches exist that promise the same guarantee, but only with some specific distributional assumptions. When those assumptions do not hold true, validity is no longer guaranteed. In contrast, CPs require only the assumption of exchangeability of data. Hence the CP is referred to as \emph{distribution-free} \citep{Lei2014}.
Central to the CP is the use of a conformity measure (CM) that, given a new example, provides a measure of how typical it is of the population. Often the CM is based on some underlying machine learning algorithm, although this is not essential.
The CP has been applied successfully in many problem domains including medical and financial, anomaly detection and network traffic applications, amongst others; see e.g. \cite{Balasubramanian2014}.

Since the accuracy of CP is essentially fixed, the optimization task is to maximize the \emph{efficiency} of the prediction sets which is some measure of the information given by the prediction set. 
There are multiple alternative measures of predictive efficiency, or conversely, predictive inefficiency\citep{Vovk2016}. For example, \cite{Colombo2020} optimize on the object fuzziness of the predictions. 
Another approach is to build CPs that minimize the mean width of prediction intervals for regression \citep{Bellotti2020} or mean prediction set size for classification. This is also the approach used for Prediction Interval Neural Networks \citep{Khosravi2011, Pearce2018}.
For clarification, Examples \ref{example1a} and \ref{example1b} give two cases of what is meant by predictive efficiency.

The CP guarantees \emph{marginal validity}: that is, across the whole population of observations, the accuracy of the CP will meet the confidence level preset by the user. However, even with marginal validity, there may be some sub-segments of the population for which accuracy is greater or less than the confidence level. 
Here are two examples demonstrating why this may be a problem.
\begin{itemize}
\item \emph{Medical example}. If diagnosing a disease, we may have a CP that gives 95\% accuracy guaranteed across all people. However, for a particular racial minority group, it gives only 85\% accuracy, whilst for the majority racial group it gives 96\% accuracy. This is clearly a problem for two reasons: (1) it makes clinical decisions less precise, especially for hospitals operating with a higher racial minority demographic, (2) it leads to medical decision support that is discriminatory.
\item \emph{Business example}. Suppose we have an automated valuation model (AVM) that gives good property price prediction intervals at 90\% confidence levels across UK. However, for the particular county of Devon its accuracy is only 75\%. This could adversely affect the business of any real estate company or mortgage provider operating in Devon who use the AVM. Worst still, the weakness of the AVM in their location may not even be known and so they will be operating under the misconception that the AVM is guaranteed to be 90\% accurate: it is in general, but not for them.
\end{itemize}
For this reason, \emph{conditional validity} has been identified as an important criterion, alongside marginal validity and predictive efficiency, for CP and indeed any algorithm generating prediction sets. That is, ideally, the validity should hold conditional on any underlying segmentation of the population. 
There are alternative forms of conditional validity depending on the type of information that accuracy is conditioned on. \cite{Vovk2013} identifies (1) \emph{label} conditional validity for conditioning on the label, (2) \emph{object} conditional validity if conditioning on the features of each example (as opposed to the label) and (3) \emph{training} conditional validity for conditioning on a property of the training set.
We can also consider combinations of these types, so if we are interested in object and label conditional validity together, we call this \emph{example} conditional validity.

\cite{Lei2014}
construct a CP which has asymptotic object conditional validity and asymptotic efficiency in the number of examples used in development (more precisely: in the calibration set), basing their CM on a conditional kernel density estimator. However, for practical purposes and in this article we are interested in finite sample results.
\cite{McCullagh2009} present provably conditionally valid prediction intervals, for the finite sample case, based specifically on linear regression, although other parametric models could be used, assuming some known probability distribution.
However, it has been shown that it is impossible to meet the criterion of conditional validity in the finite sample case for non-trivial problems in the general case \citep{Vovk2013, Lei2014}. Both papers suggest label-conditioned CPs for classification as a way to address this problem. More generally, a conditional CP can be used that allows the user to specify a taxonomy of data into finite categories and then the CP is valid within those pre-defined categories \citep{Vovk2014}. However, these approaches are problematic since (1) they limit the conditional validity to just some finite pre-defined categories and do not apply generally to \emph{any} taxonomy, and (2) the effectiveness of the CP is reduced with the more categories included in the taxonomy, since the conditional CP
essentially relies on a separate CP for each segment of data, limiting these approaches for practical use with large taxonomies.

These practical limitations with conditional validity have led \cite{Barber2020} to proprose relaxing the requirement for conditional validity and to explore using an \emph{approximation} to conditional validity. 
They set up a framework called \emph{distribution-free approximate conditional coverage} meaning that conditional validity should be true for at least some pre-defined proportion of the population, which is similar to the PAC-type conditional validity given by \cite{Vovk2014}. However, in the general case, they find that it is also impossible to meet this broad target. They make further progress proposing \emph{restricted conditional coverage} to provide local conditional validity in the neighbourhood (ball) around some point. This is a promising approach and they show that CP satisfies this condition, but only for the case of CP for regression with the standard CM. Additionally, they identify computational difficulties with this method that require further research.
The development of the normalized CM for regression \citep{Papadopoulos2002InductiveRegression} is an approach which can be seen as addressing this problem since the normalization factor allows more difficult prediction cases to be given wider prediction intervals, hence increasing their probability of being correct, and vice versa for easier prediction cases. However, it is a heuristic approach since it does not directly optimize for approximation to conditional validity and there has been little investigation of its effectiveness for approximation to conditional validity.

It may be supposed that pursuing the goal of maximizing predictive efficiency may align with conditional validity, in some sense. 
However, Example \ref{example2} gives a simple case demonstrating predictive efficiency is not consistent with conditional validity.
In this paper, we focus on object conditional validity. The medical and business examples given above are both examples of this. We demonstrate that lack of object conditional validity is a genuine problem for CP through experiments with three data sets and alternative CPs; and, secondly, propose a new \emph{iterative feedback-adjusted conformity measure} (IFACM) algorithm that can be used as part of an inductive conformal predictor (ICP) to achieve approximation to object conditional validity (AOCV), although in a different sense to \cite{Barber2020}. We provide some theory and experimental results. The methodology is based on ICP since ICP is a practical version of CP appropriate for batch machine learning.
The proposed IFACM algorithm works from the proposition that (1) the AOCV of an ICP can be estimated using a meta-model, and (2) this meta-model can then be used to adjust the CM thus forming a new ICP which is able to achieve better AOCV on the same problem. This process can be repeated until no improvement in AOCV can be acheived. Finally, this algorithm will output a CM formed as layers of adjustments to some base CM. The algorithm is similar in approach to that of boosting algorithms that are used to reduce errors on point predictions through layers of simple predictive models \citep{Freund99ashort}.

The remainder of this article is organized as follows. Section \ref{methodology} introduces CP and the proposed IFACM algorithm, Section \ref{experiments} presents some experimental results for both regression and classification, demonstrating the effectiveness of the proposed algorithm to achieve AOCV and conclusions are given in Section \ref{conclusion}. The Appendix provides supporting theoretical results. 
Abbreviations used in this article are listed in Table \ref{tab:abbreviations} for the convenience of the reader.
\begin{table}[ht]
\centering
\begin{tabular}{ll}
  \hline
AOCV & Approximation to object conditional validity \\
CM & Conformity measure \\
CP & Conformal predictor \\
DCV & Deviation from conditional validity \\
ICP & Inductive conformal predictor \\
IFACM & Iterative feedback-adjusted conformity measure \\
\hline
\end{tabular}
\caption{Abbreviations used in this article.} 
\label{tab:abbreviations}
\end{table}

\begin{framed}
\begin{example} \label{example1a} 
Suppose a disease subtype needs to be diagnosed and it must be one of $A, B$ or $C$.
If the predictor outputs the prediction set $\{A,B,C\}$ then it provides no new information at all, since the prediction set just states that the subtype is any of those available. 
The most informative prediction is the prediction singleton set, say $\{B\}$, since this isolates one single case.
The prediction set $\{A,B\}$ offers some information (by excluding $C$) but is not as informative as the singleton set.
Hence this example suggests that cardinality of the prediction set is a good measure of predictive inefficiency.
\end{example}
\begin{example} \label{example1b} 
Suppose property price is predicted and a prediction interval $[230,534]$ (in '000 GBP) is given. Then this range is rather broad and may not be particularly useful to a real estate agent. If the prediction algorithm is improved to give prediction interval $[240, 350]$ then this may prove an informatve and useful prediction. 
Hence, this example suggests that width of prediction interval would be a good measure of predictive inefficiency. For these particular examples, the predictive inefficiency would be $304$ and $110$ respectively.
\end{example}
\end{framed}

\begin{framed}
\begin{example} \label{example2}
A simple example is constructed to demonstrate that finding the most efficient prediction intervals, on average, whilst maintaining marginal validity, leads to conditionally invalid prediction intervals.

Suppose we have one predictor variable $X \in \{1,2\}$ with $\mathbb{P}(X=1)=\frac{1}{2}$, and a label that is uniformly distributed, $Y \sim \textrm{U}(0, X)$.
A simple predictor makes prediction intervals $R=[0,a]$ if $X=1$ and $R=[0,b]$ if $X=2$, so parameters $a$ and $b$ need to be computed.
A significance level $\varepsilon < \frac{1}{2}$ is preset for marginal validity, i.e. $\mathbb{P}(Y \in R) =  1-\varepsilon$.

By the law of total probability,
$$\begin{array}{ll}
\mathbb{P}(Y \in R)  
&= \mathbb{P}(Y \in [0,a]\; | X=1) \mathbb{P}(X=1) +  \mathbb{P}(Y \in [0,b]\; | X=2) \mathbb{P}(X=2) \\
&= \mathbb{P}(Y \in [0,a] \;| X=1)\times \frac{1}{2} +  \mathbb{P}(Y \in [0,b]\; | X=2)\times \frac{1}{2} \\
& = \frac{1}{2} a + \frac{1}{4} b = 1-\varepsilon.
\end{array} $$
Measure inefficiency as width of the prediction interval, then minimizing mean inefficiency is minimizing 
$$\mathbb{E}(|R|)=\mathbb{E}(|R| \;  | X=1) \mathbb{P}(X=1) + \mathbb{E}(|R| \;  | X=2) \mathbb{P}(X=2)
= \frac{1}{2} a + \frac{1}{2}b.$$ 
With the marginal validity constraint above, this is a simple linear programming problem with the solution,
$$a=1, \quad b=2-4\varepsilon,$$
noticing that since $\varepsilon < \frac{1}{2}$, $b > 0$ 
(an alternative is to use the square of prediction interval width as an inefficiency measure, in which case $\frac{1}{2} a^2 + \frac{1}{2}b^2$ is minimized; however, this leads to the same solution).
This then leads to the two conditional probabilities of accuracy:
$$ \mathbb{P}(Y \in [0,a] \;| X=1) = 1, \quad \mathbb{P}(Y \in [0,b] \;| X=2)=1-2\varepsilon$$
which both deviate from the confidence level $1-\varepsilon$, hence the solution does not exhibit conditional validity.


It is possible to enforce object conditional validity for this simple example by setting $a=1-\varepsilon$ and $b=2(1-\varepsilon)$ so that 
$$ \mathbb{P}(Y \in [0,a] \;| X=1) = \mathbb{P}(Y \in [0,b] \;| X=2)=1-\varepsilon$$
However, mean inefficiency then increases from $\frac{3}{2}-2\varepsilon$ to $\frac{3}{2}(1-\varepsilon)$, a difference of $\frac{1}{2}\varepsilon$ more than optimal inefficiency.
Therefore enforcing object conditional validity leads to a reduction in predictive efficiency.
\end{example}
\end{framed}

\section{Methodology} \label{methodology}

We first introduce ICP and marginal validity, then discuss AOCV before introducing the IFACM algorithm. 
Finally we show that IFACM for regression is consistent with the normalized CM.

\subsection{Inductive Conformal Predictors}

We set up the ICP framework following \cite{Vovk2005AlgorithmicWorld}.
\begin{itemize}
\item Let $\bs{z}= (\bs{x}, y)$ denote examples from the same unknown distribution $P$ such that $\bs{x}$ is a vector of $m$ predictor variables $\bs{x} \in \mathbb{R}^m$ and label $y \in \mathbb{Y}$ for some set of labels $\mathbb{Y}$.
\item Let $\bs{z}_1, \cdots, \bs{z}_l$ be a particular sequence of examples $\bs{z}_j = (\bs{x}_j, y_j)$ from $P$.
\item Let $1$ to $k$ index the training data set and $k+1$ to $l$ index the calibration set,
for $1<k<l$.
\item A conformity measure (CM) is any computable function 
$$A(\bs{z}) = \mathcal{A}(\bs{z}_1, \cdots, \bs{z}_k,  \bs{z} )$$
such that $ \mathcal{A}$ is exchangeable: 
ie $ \mathcal{A}( \bs{z}_1, \cdots,  \bs{z}_k, \bs{z} ) 
=  \mathcal{A}(\bs{z}_{\pi(1)}, \cdots,  \bs{z}_{\pi(k)},  \bs{z} ) $ 
for all permutations $\pi$ of $1, \cdots, k$.
For convenience, we will sometimes use $A(\bs{x}, y)$ to denote $A(\bs{z})$.
\item Let $\varepsilon$ be a preset significance level, so $1-\varepsilon$ is the confidence level for predictions.
\end{itemize}
We typically consider a CM having the form $A(\bs{z}) = \mathcal{A}^p(\theta,  \bs{z} )$ for some parameters $\theta = M(\bs{z}_1, \cdots,  \bs{z}_k)$
where $M$ is intended to be a model structure within which $\theta$ are estimated based on data $\bs{z}_1, \cdots,  \bs{z}_k$. For example, for regression, $M$ could be ordinary least squares (OLS) linear regression and $\theta$, the vector of coefficients in the OLS linear regression model; or $M$ could be an artificial neural network with $\theta$ the vector of weights in the network.

The ICP is defined as the prediction algorithm that gives the prediction set at significance level $\varepsilon$ for a new example $\bs{x}$, based on the calibration set of examples $k+1$ to $l$ as
\begin{equation} \label{eq:predinterval1}
\Gamma^{\varepsilon}(\bs{x}; A) 
	= \left\{y \in  \mathbb{Y} :  \sum_{j=k+1}^l {\mathbb{I} \left[ A(\bs{x},y) \ge A(\bs{z}_j) \right]} + 1 
  >  \varepsilon (l-k+1) \right\}
\end{equation}
where $\mathbb{I}$ is the usual indicator function. This prediction set is dependent on the CM used so this is made explicit on the left-hand side of the expression. Assuming only that the calibration data set, $\bs{z}_{k+1}, \cdots, \bs{z}_l$, and any new test or operational data sets are exchangeable, ICP predictions on the new data are marginally valid:
\begin{equation} \label{eq:validity}
\mathbb{P}(y \in \Gamma^{\varepsilon}(\bs{x}; A) ) = 1-\varepsilon.
\end{equation}
for each new example $(\bs{x},y)$
\footnote{Strictly speaking, exact validity is only true for the smoothed CP \citep{Vovk2005AlgorithmicWorld}.
In the case when CM generates ties, we would have conservative validity: $\mathbb{P}(y \in \Gamma^{\varepsilon}(\bs{x}) ) \ge 1-\varepsilon$.  However, for practical purposes with sufficient granularity in the CM, exact validity is approximately the case for ICP.}. 
When the label set $\mathbb{Y}$ is finite, this is ICP for classification. 
When $\mathbb{Y}=\mathbb{R}$, this is ICP for regression.
 For these two cases, we can consider two particular CMs that are common in the literature.
\begin{description}
\item[Scoring CM]
A scoring function $f:\mathbb{R}^m \times \mathbb{Y} \to \mathbb{R}$ for which a higher output of $f(\bs{x}, y)$ means that label $y$ is (in some sense) more likely to be associated with example $\bs{x}$, then $f$ itself can be used directly as a CM.
An example of this is multinomial logistic regression for classification which estimates coefficient vectors $\beta_y$ for each possible label $y \in \mathbb{Y}$ such that $f(\bs{x}, y) = \beta_y \cdot \bs{x}$ is the log-odds of $y$ conditional on $\bs{x}$. We will use the multinomial logistic regression scoring function in the later experimental section.
\item[Normalized CM]
In the case of regression, a typical choice is the normalized nonconformity measure:
\begin{equation} \label{eq:ncm}
N(\bs{x}, y) = \frac{|y-\hat{y}|}{\hat\sigma}
\end{equation}
where $\hat{y}$ is a point estimate based on a regression model of $y$ and $\hat\sigma >0$ is some estimate of uncertainty in the prediction given by the model \citep{Papadopoulos2002InductiveRegression}. Both can be computed from models estimated from training data, $\bs{z}_1, \cdots,  \bs{z}_k$.
For later experiments in this article, OLS linear regression of $y$ is used for the numerator and a OLS linear regression of the log of absolute residual is used for $\sigma$ following \cite{Papadopoulos2002InductiveRegression}.
For consistency, we use CM for this study and a normalized CM can be constructed as the reciprocal of the normalized nonconformity measure $A_{\textrm{N}} (\bs{x}, y) = 1 / N(\bs{x}, y)$.
For the normalized nonconformity measure
\begin{equation} \label{eq:predinterval_ncm}
\Gamma^{\varepsilon}(\bs{x};  A_{\textrm{N}}) 
	= [ \hat{y} - \hat\sigma q, \; \hat{y} + \hat\sigma q ]
\end{equation}
thus forming a prediction interval, where $q$ is the $\varepsilon$-quantile of the empirical distribution of CM values within the calibration set.
In the special case when $\hat\sigma=1$, this gives the \emph{standard CM}. As can be seen from (\ref{eq:predinterval_ncm}), the standard CM always gives the same fixed width $2q$ for prediction intervals.
The introduction of the normalizing denominator allows for variation in the prediction interval width and hence form a CP that has varied prediction interval widths according to the difficulty of the observation, with observations that are more difficult to predict having wider prediction intervals.
\end{description}

\subsection{Approximation to Object Conditional Validity}

\newcommand\proba{\mathbb{P}(y \in \Gamma^{\varepsilon}(\bs{x}; A) \; | \; \bs{x})}

As discussed earlier, it would be valuable to achieve object conditional validity given by
$$\proba=1-\varepsilon$$ for all $\bs{x} \in \mathbb{R}^m$,
but, in the nontrivial case, this is impossible to achieve for CPs as shown by\cite{Vovk2013} and \cite{Lei2014}.
Instead we consider relaxing the constraint and allow for AOCV given broadly by
$$\proba \approx 1-\varepsilon.$$
This requirement can be made more precise by expressing it as minimizing the square difference across the population,
\begin{equation} \label{eq:expdiff}
\mathbb{E}_{\bs{x}} \left[ {\left( \proba - (1-\epsilon ) \right)}^2\right].
\end{equation}
This approach to AOCV is quite different to \cite{Barber2020} who, instead, consider proportions of the population that are exactly conditionally valid as an approximation method.
After conformal predictions are made on a data set indexed by some index set $S$, (\ref{eq:expdiff}) is estimated as
$$\frac{1}{|S|} \sum_{j \in S} {{( p( \bs{z}_j; A) - (1-\epsilon ))}^2}$$
where
\begin{equation} \label{eq:probest}
p( (\bs{x}, y); A) = \hat{\mathbb{P}}(y \in \Gamma^{\varepsilon}(\bs{x};A) \; | \; \bs{x}) 
\end{equation}
is some probability estimator.
The probability estimator  can be logistic regression, $k$-nearest neighbours ($k$-NN) or any other probabilistic model taking training data 
$(\bs{x}_j, c_j) \forall j \in S$ where 
$$c_j =  \mathbb{I} \left[ y_j \in \Gamma^{\varepsilon}(\bs{x}_j; A) \right]$$
indicates whether the prediction set for example $j$ is correct.
Therefore, this measure can be used to quantify AOCV. We take the square root of this estimate, so that the measure is on the same scale as the conditional probability. Hence the root mean square error gives \emph{deviation from conditional validity} (DCV),
$$\textrm{DCV}(A)=\sqrt{\frac{1}{|S|} \sum_{j \in S} {{( p( \bs{z}_j; A) - (1-\epsilon ))}^2}}. $$
If an ICP is conditionally valid then $\textrm{DCV}(A) \approx 0$.
The closer it is to zero, the better the AOCV exhibited by the ICP.  Essentially $p( \bs{z}; A) \approx 1-\varepsilon$ indicating an uninformative model. For example, using logistic regression for $p$, the coefficient estimates will be close to zero and statistically insignificant.

On the other hand, a low value of DCV does not \emph{guarantee} conditional validity, since it only reflects the model structure that is used to construct the probability estimate, such as logistic regression, and not all possible conditionalities in the data. However, it will be a good indicator of AOCV for practical purposes.

\subsection{Iterative Feedback-adjusted Conformity Measure}

Consider constructing CMs that seek to minimize $\textrm{DCV}(A)$.
Our proposal is to extend an existing base CM $A$ which may have a high DCV with another as follows:
$$\asgg{\bs{z}} = A(\bs{z}) + \gamma U(\bs{z}; A)$$
where $\gamma > 0$ is a control parameter, $U$ is an update function based on the discrepency between estimated probability given $A$ and confidence level, with the properties,
\begin{itemize}
\item $U(\bs{z}; A) \ge 0$ if $p( \bs{z}; A) < 1-\varepsilon$;
\item $U(\bs{z}; A) = 0$ if $p( \bs{z}; A)= 1-\varepsilon$;
\item $U(\bs{z}; A) \le 0$ if $p( \bs{z}; A) > 1-\varepsilon$.
\end{itemize}
where the underlying estimator $p$ is exchangeable, i.e. the estimates are invariant to the order of the training examples. This ensures that $A^*_{\gamma}$ is also a CM. This exchangeability condition is true of most probabilistic estimators and, in particular, is true of logistic regression and $k$-NN where the order of training examples is irrelevant.
Hence, if the training data for $p$  is independent of calibration and test data, then the use of $A^*_{\gamma}$ in an ICP guarantees marginal validity (\ref{eq:validity}).

The inituition for the update function is that if the conditional probability of being correct is too low, then we increase the CM so that the correct label has more chance of being in the prediction interval (\ref{eq:predinterval1}), hence increasing the conditional probability of being correct if ICP is rerun with the updated CM, and vice versa, for conditional probabilities that are too high, the CM value is lowered following the update, thus making it less likely that the prediction is correct. 

A specific update function considered in this study is
$$U^* (\bs{z}; A, \delta) = \left\{
\begin{array}{ll}
+1 &\textrm{if }p( \bs{z}; A) < (1-\varepsilon)-\delta \\
-1 &\textrm{if } p( \bs{z}; A) > (1-\varepsilon)+\delta \\
0 &\textrm{otherwise}
\end{array} \right.$$
where $\delta \ge 0$ is a sensitivity parameter. Notice that $U^*$ is not dependent on $y$ but this is not a problem, since $A$ is expected to provide discrimatory power for the ICP, whilst $U^*$ just provides an adjustment to lead the ICP to achieve better AOCV. 
Ideally we would like to prove that adjusting by $U^*$ will improve AOCV in some way: perhaps showing that DCV is reduced. However, this seems difficult to achieve, at least in the general case, and remains an area for further research. However, it is possible to prove that applying $U^*$ adjustment to the CM does improve the average log-odds of being correct for those observations whose outcome (i.e. $c_j$) is switched by the adjustment, so that it is closer to the target confidence level, in the case when logistic regression is used as the probability estimator $p$. Technical details are provided in Appendix A.

The updated CM $A^*_{\gamma}$ is optimized with respect to $\gamma$ and $\delta$ using the training data set with the goal to minimize the $\textrm{DCV}$ measure. There is no gradient that can be used for optimization but with only two parameters to minimize across, Nelder-Mead optimization \citep{Nelder1965} is adequate for this task. Since both parameters are greater or equal to zero, the optimization is performed over the logarithm of the parameters.

The $U^*$ update function is rather simplistic and one application of an update may provide just a small improvement in $\textrm{DCV}$. However, since the update $A^*$ is itself a CM, the application of the update function can be applied iteratively, minimizing DCV at each iteration, until it cannot be minimized any further. Representing the CM recursively:
$$A^{*[i]}(\bs{z})= \left\{
\begin{array}{ll}
A(\bs{z}) & \textrm{if } i=0, \\
A^{*[i-1]}(\bs{z}) + \gamma_i U^*(\bs{z}; A^{*[i-1]}, \delta_i) & \textrm{if } i>0
\end{array} \right.$$
for parameters $\gamma_i>0$ and $\delta_i \ge 0$ that are computed at each iteration to minimize DCV. 
Importantly, the probability estimator (\ref{eq:probest}) is re-estimated at each iteration based on the ICP with the current version of $A^{*[i]}$.
Then the output after $n_u$ iterations is the IFACM expressed as the weighted sum,
\begin{equation} \label{eq:ifacm}
A^{**}(\bs{z}) = A(\bs{z}) + \sum_{i=1}^{n_u} \gamma_i U^*(\bs{z}; A^{*[i-1]}, \delta_i). 
\end{equation}
This use of iterative modifications to the CM based on feedback from previous stages of the ICP performance mirrors the method of boosting that is used to reduce errors iteratively by giving greater weight to examples with errors from previous iterations in each model build at the next iteration. 
We would expect that the process of adjusting the CM by iterative updates will lead to reduced DCV, but on the other hand, that there will be a trade-off in an increase in predictive inefficiency. Hence, to mitigate against this the optimization task is modified to minimize at each iteration $i$,
\begin{equation} \label{eq:opt1}
\textrm{DCV}(A^{*[i]})+ C (I( A^{*[i]}) - I(A) ) \mathbb{I} \left[I(A^{*[i]})  > I(A) \right] 
\end{equation}
where $I$ is an inefficiency measure for predictions made using the CM on index set $S$  and $C \ge 0$ is a control hyperparameter. 
There are several alternative measures of prediction inefficiency \citep{Vovk2016}. In this study, the mean size of prediction sets is used,
\begin{equation} \label{eq:mean_width}
I(A')= \frac{1}{|S|} \sum_{j\in S} {\left| \Gamma^\varepsilon (\bs{x}_j; A') \right|}.
\end{equation}
For classification, the size is the cardinality of the prediction set. For regression, using normalized CM, the size is the width of the prediction interval.

The method is introduced with the abstract set $S$. However, in practice, this would typically be the training data; i.e. $S=\{1,\cdots, k\}$, in order to conserve data. The consequence of this is that the ICP constructed at each iteration of the algorithm is \emph{not marginally valid} since the training, calibration and independent test sets all coincide, leading to overfitting. However, ultimately this is not a problem since the IFACM that is computed is subsequently an input to a proper ICP that does have an independent calibration set and independent testing (or operational) data. Algorithms \ref{alg1} and \ref{alg2} show how to compute the IFACM that can then be used within an external ICP with an independent calibration data set. 

\begin{algorithm}[ht]
\SetAlgoLined
\SetKwInOut{Input}{input}\SetKwInOut{Output}{output}
\Input{\emph{Train} = training data set, \\ 
            \emph{CL} = confidence level, \\
            $A$ = conformity measure.
}
  Compute prediction intervals \emph{PI} for all observations in \emph{Train} using ICP with $A$, \emph{CL} and \emph{Train} as the calibration data set\;
  Construct a probabilitic model \emph{M} of the correctness of \emph{PI} on \emph{Train} \;
  Compute $DCV$ of \emph{M} predicting on \emph{Train}, relative to \emph{CL} \;
  $W \leftarrow$ mean width of \emph{PI}, using (\ref{eq:mean_width}) \;
\Output{$DCV, W$}
 \caption{Run ICP on training data to measure DCV}
\label{alg1}
\end{algorithm}

\begin{algorithm}[ht]
\SetAlgoLined
\SetKwInOut{Input}{input}\SetKwInOut{Output}{output}
\Input{\emph{Train} = training data set, \\ 
            \emph{CL} = confidence level, \\
            $C$ = control parameter in equation (\ref{eq:opt1}), \\
            $B$ = Base conformity measure. 
}
  $(DCV_1, W_0) \leftarrow$ Algorithm \ref{alg1} with input: \emph{Train}, \emph{CL}, \emph{B}\;
  $A \leftarrow B$ \;
  $i \leftarrow 1$ \;

 \Repeat{$DCV_2 \ge DCV_1$}{
  Run Algorithm \ref{alg1} multiple times with input:  \emph{Train}, \emph{CL},  $A + \gamma_i U^*(.;A,\delta_i)$ to minimize $DCV+C(W-W_0)\mathbb{I}(W>W_0)$  with respect to $\gamma_i, \delta_i$ using the Nelder-Mead algorithm
(\emph{i.e. minimize (\ref{eq:opt1}) with respect to $\gamma_i, \delta_i$}) \; 
$DCV_2 \leftarrow DCV$ from the minimization step above \;
\If{$DCV_2 < DCV_1$} { 
   $DCV_1 \leftarrow DCV_2$ \;
   $A \leftarrow A + \gamma_i U^*(.;A,\delta_i)$ \;
   $ i \leftarrow i+1$ 
  }
}
$n_u \leftarrow i$ \;
\Output{$A$}
 \caption{Iterative feedback-adjusted conformity measure (IFACM)}
\label{alg2}
\end{algorithm}

\subsection{IFACM and Normalized CM}

The CM $A^{**}$ is consistent with the normalized CM for regression. To see this, first observe that applying a monotonically increasing transform to a CM does not change the behaviour of the ICP since it is the rank ordering of CM that matters, as can be seen in (\ref{eq:predinterval1}).  Therefore, taking base $A(\bs{x}, y) = \log \sigma - \log{|y-\hat{y}|}$ and applying the exponential function to (\ref{eq:ifacm}) gives
$$\exp A^{**}(\bs{x}, y) = \frac{\hat\sigma}{|y-\hat{y}|}
 \exp \left[ \sum_{i=1}^{n_u} \gamma_i U^*(\bs{x}, y; A, \delta_i) \right].$$
Since the exponential function is monotonically increasing, this CM will give the same ICP as (\ref{eq:ifacm}), except in the transformed version it can be seen that this is the normalized CM with an extra product update. Indeed, the update functions can be interpreted as a further normalizing factor in the CM. Therefore from (\ref{eq:predinterval_ncm}), the prediction interval yielded by this CM for regression is
\begin{equation} \label{eq:predinterval_ncm2}
\Gamma^{\varepsilon}(\bs{x};  A^{**}) = \left[ \hat{y} - \hat\sigma' q, \; \hat{y} + \hat\sigma' q \right]
\end{equation}
where
$$\hat\sigma' = \hat\sigma \exp \left[ \sum_{i=1}^{n_u} \gamma_i U^*(\bs{x}, y; A, \delta_i) \right].$$


\section{Experimental settings and results} \label{experiments}

Experiments were conducted with three data sets listed in Table \ref{tab:datasets}.
Data set ``covtype'' has outcome identifying tree cover type for different forest locations. It is used in this study since it is a large multi-class classification problem. 
Data set ``GPU'' provides GPU performance data for different GPU settings, outcome is average performance time and it is a large regression data set \citep{GPU2015} .
Data set ``KC'' (King's County) contains US housing data and the outcome variable is property sale price. It is a smaller regression problem but is interesting for this study since the normalized CM gives improved performance over the standard CM, which was not the case with the GPU regression problem. Both ``covtype'' and ``GPU'' are available from the UCI Machine Learning data repository \citep{Frank2010UCIRepository} and ``KC'' is available from the Kaggle website (\texttt{www.kaggle.com}).
\begin{table}[ht]
\centering
\begin{tabular}{l r l r r r }
  \hline
Data set & \#var &  Outcome (\#labels) & \#Train & \#Cal & \#Test  \\ 
  \hline
covtype  & 54 & Classification (7) & 100000& 100000& 100000 \\
   \hline
        GPU & 14 & Regression & 20000 & 20000 & 20000 \\ \hline
        KC & 24 & Regression & 10000 & 5000 & 6613 \\ \hline
\end{tabular}
\caption{Data sets. 
\#var = number of predictor variables, 
\#labels = number of unique class labels for classification,
\#Train, \#Cal, \#Test are number of observations in the training, calibration and test sets, respectively.} 
\label{tab:datasets}
\end{table}

\begin{figure} 
\centering
\includegraphics[scale=0.3] {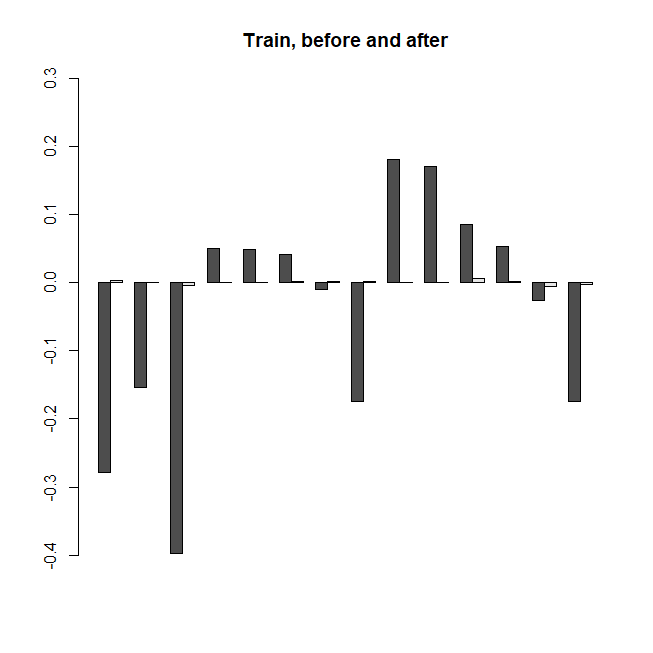}
\includegraphics[scale=0.3] {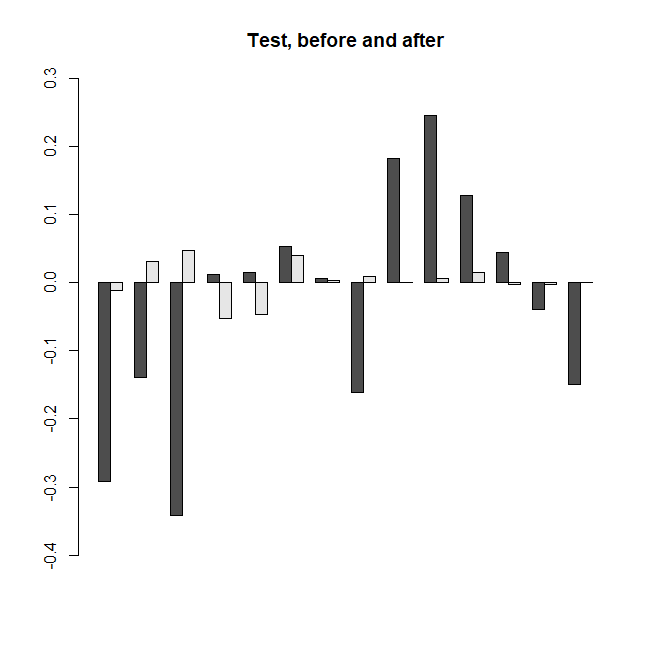}
 \caption{Change in coefficient estimates in probability estimator $p$ before (dark grey) and after (light grey) applying the IFACM algorithm on ``GPU''  training data (left) and test data (right).} 
\label{fig:GPU-coeff-change}
\end{figure}

Parameter settings are as follows:-
\begin{itemize}
\item \emph{Confidence level}. 
For all experiments, confidence level is set to make the problem sufficiently hard that predictive efficiency becomes a challenge. For ``covtype'' this is $0.95$ and for ``GPU'' and ``KC'' it is $0.9$. Additionally a confidence level of $0.99$ is also used for ``GPU'' to explore the consequences of high confidence levels. 
\item \emph{Base CM}. 
For classification, the multinomial scoring CM is used as base CM, whereas normalized CM is used for the regression problems. Additionally standard CM for regression is used for ``KC'' to explore if normalized CM improves on standard CM in terms of AOCV. Both use a base OLS linear regression model for point estimates of $\hat{y}$ and normalized CM also uses OLS linear regression for estimates of $\log\hat\sigma$ for (\ref{eq:ncm}).
\item \emph{Probability estimator}.
Logistic regression is used as the probability estimator (\ref{eq:probest}).
\item \emph{$C$: penalty hyperparameter in (\ref{eq:opt1})}.
For ``covtype'', it was possible to achieve good results with $C=0$. However, for the other data sets, $C=0$ leads to huge predictive inefficiency, hence a value of $C=0.5$ is used instead. For contrast, $C=5$ is also used.
\item \emph{Nelder-Mead optimization}.
Low starting values $\log \gamma_i=-5, \; \log\delta_i=-5$ were used when minimizing (\ref{eq:opt1}).
\end{itemize}
Results are shown in Table \ref{tab:results}. The main points that we find are as follows.
\begin{itemize}
\item For all experiments, the IFACM algorithm successfully reduces DCV on both the training and test data sets, whilst maintaining marginal validity (i.e. Acc. $\approx$ CL).
\item As expected, the reduction in DCV comes at the cost of increase in predictive inefficiency and this is controlled by the hyperparameter $C$. However, the impact of this trade-off is different for each data set. For ``covtype'', inefficiency increases from 1.78 to 1.84 for the best improvement in DCV (when $C=0$, lines 3--4), whereas taking $C=5$ will yield better inefficiency (1.80) but the DCV more than doubles, so the impact of the trade-off between DCV and inefficiency is high for this data set. 
However, for ``GPU'', a low DCV is measured with minimal increase in inefficiency when $C$ is low (lines 11--12).
Surprisingly, when a high confidence level is used, IFACM yields lower inefficiency even with low DCV (lines 15--18). Whether this is a general result requires further investigation, but at the very least it demonstrates that for ``GPU'', the impact of applying IFACM on inefficiency is minimal.
\item In many cases, DCV is much lower on the training data set compared to the test set. This is especially the case when low values of $C$ are used (compare line 11 to 12 and 25 to 26). This suggests that in some situations, the IFACM algorithm gives rise to a large degree of overfit for the goal of minimizing DCV. For these data sets, it does not give rise to a significant problem with performance on the test data; however, it is worth some further investigation.
\item Lines 19--22 demonstrate that for the ``KC'' regression problem, using NCM gives lower predictive inefficiency than the standard CM which confirms what we expect from previous work (e.g. \citep{Papadopoulos2002InductiveRegression}), but also interestingly the normalized CM is able to reduce DCV on training and test data (lines 21--22), which is expected by construction of the normalized CM.
Nevertheless, we see that applying IFACM with the normalized CM as base CM allows for even further reduction in DCV. Using $C=5$, IFACM is able to achieve this with only a small increase in inefficiency (lines 29--30).
Interestingly, applying IFACM on the standard CM gives improvement over standard CM, but is not as good as IFACM applied to normalized CM in terms  of either inefficiency or DCV measures (compare lines 27--28 to 29--30). This is important since it shows that IFACM cannot simply replace a good base CM (i.e. the normalized CM) but is a supplement to an existing base CM, for the purpose of achieving AOCV.
\end{itemize}
\begin{table}[ht]
\centering
\begin{tabular}{r | l l l l l | lll}
  \hline
\# & Data set & Segment &  Algorithm & CL & $C$ & Acc. & Ineff. & DCV  \\ 
  \hline
1& covtype  
  & training & Base ICP & 0.95 & - & 0.951 & 1.78 & 0.0396 \\
2&  & test       & Base ICP & 0.95 & - & 0.950 & 1.78 & 0.0395 \\
3& & training & IFACM      & 0.95 & 0 & 0.951 & 1.84 & 0.00345 \\
4&  & test       & IFACM      & 0.95 & 0 & 0.950 & 1.84 & 0.00705 \\
5&  & training & IFACM      & 0.95 & 1 & 0.950 & 1.82 & 0.00953 \\
6&  & test       & IFACM      & 0.95 & 1 & 0.949 & 1.83 & 0.0101 \\
7&  & training & IFACM      & 0.95 & 5 & 0.950 & 1.80 & 0.0193 \\
8&  & test       & IFACM      & 0.95 & 5 & 0.949 & 1.80 & 0.0185 \\   
 \hline
9& GPU 
 & training & Base ICP / NCM & 0.9 & - & 0.901 & 2.20 & 0.0497 \\
10&  & test       & Base ICP / NCM & 0.9 & - & 0.898 & 2.20 & 0.0503 \\
11& & training & IFACM / NCM & 0.9 & 0.5 & 0.901 & 2.23 & 0.000949 \\
12&  & test       & IFACM / NCM & 0.9 & 0.5 & 0.898 & 2.23 & 0.00725 \\
13& & training & IFACM / NCM & 0.9 & 5 & 0.901 & 2.20 & 0.00497 \\
14&  & test       & IFACM / NCM & 0.9 & 5 & 0.901 & 2.20 & 0.00954 \\
\hline
 15 & GPU 
  & training & Base ICP / NCM & 0.99 & - & 0.991 & 3.38 & 0.0162 \\
16&  & test       & Base ICP / NCM & 0.99 & - & 0.991 & 3.38 & 0.0181 \\
17& & training & IFACM / NCM & 0.99 & 0.5 & 0.991 & 3.27 & 0.00117 \\
18&  & test       & IFACM / NCM & 0.99 & 0.5 & 0.991 & 3.28 & 0.00260 \\
\hline
19 & KC 
  & training & Base ICP / SCM & 0.9 & - & 0.908 & 1.58 & 0.121 \\
20&  & test      & Base ICP / SCM & 0.9 & - & 0.900 & 1.58 & 0.135 \\
21& & training & Base ICP / NCM & 0.9 & - & 0.903 & 1.49 & 0.0549 \\
22&  & test      & Base ICP / NCM & 0.9 & - & 0.900 & 1.49 & 0.0587 \\
23&  & training & IFACM / SCM & 0.9 & 0.5 & 0.900 & 1.70 & 0.0359 \\
24&  & test      & IFACM / SCM & 0.9 & 0.5 & 0.900 & 1.69 & 0.0316 \\
25&  & training &IFACM / NCM & 0.9 & 0.5 & 0.903 & 1.55 & 0.00200 \\
26&  & test      &IFACM / NCM & 0.9 & 0.5 & 0.900 & 1.55 & 0.0181 \\
27&  & training & IFACM / SCM & 0.9 & 5 & 0.901 & 1.60 & 0.0475 \\
28&  & test      & IFACM / SCM & 0.9 & 5 & 0.900 & 1.59 & 0.0455 \\
29&  & training &IFACM / NCM & 0.9 & 5 & 0.902 & 1.52 & 0.0162 \\
30&  & test      &IFACM / NCM & 0.9 & 5 & 0.900 & 1.52 & 0.0199 \\
\hline
\end{tabular}
\caption{Experimental results. 
SCM=standard CM,
NCM=normalized CM,
CL=confidence level $1-\varepsilon$, 
$C$=penalty hyperparameter in (\ref{eq:opt1}),
Acc.=accuracy,
Ineff.=mean prediction interval width.} 
\label{tab:results}
\end{table}
We can look in further detail at the behaviour of the IFACM algorithm. 
\begin{itemize}
\item Figure \ref{fig:GPU-coeff-change} shows the coefficient estimates for each of the variables in ``GPU'' using the logistic regression probability estimator $p$ in (\ref{eq:probest}), before and after IFACM is applied, corresponding to lines 10 and 12 of Table \ref{tab:results} respectively. Since data has been standardized, the magnitude of the coefficients are comparable. The left graph shows the result on training data: in this case, the reduction in magnitude of coefficients is large, leading to an uninformative model after IFACM (i.e. coefficients $\approx 0$)
\footnote{For this problem, notice that we are in the peculiar position of wanting to achieve the \emph{worst} model in the sense of weak model fit, so that there is no relationship between whether a prediction interval for an object $\bs{x}$ is correct and $\bs{x}$ itself.}. The right graph shows coefficients on test data: overall, there is a reduction in magnitude, but it is not as pronounced as with training data and some coefficients actually become larger in magnitude. This demonstrates there is overfitting with this algorithm, i.e. it does not perform so well on test data as training, in terms of AOCV, corroborating the results for DCV given in Table \ref{tab:results}.
\item Table \ref{tab:gamma_delta} shows the sequence of values of parameters $\gamma_i$ and $\delta_i$ estimated in several experiments. They show that the number of iterations of the IFACM algorithm is not large, with maximum $5$ in our experiments. Also, the number of iterations goes down as $C$ increases, showing that greater control of inefficiency means less precision with the feedback adjustment. The sequence of values of $\gamma_i$ decreases with each iteration, as the magnitude of the adjustment at each iteration is reduced. Finally, however, we notice no particular pattern with the sequences of $\delta_i$ values.
\item Although DCV reported in Table \ref{tab:results} measures average deviation from conditional validity across the whole data set, to convince ourselves that the method is effective, it is also valuable to show accuracy for different sub-segments of the data. Therefore, as an illustration, Table \ref{tab:segments} shows this for several segments in the test data sets. These segments are chosen as deviating greatly from the confidence level when the Base ICP is used. The results show that for each of these segments, using IFACM makes a considerable difference, giving accuracies much closer to the target confidence level, within each segment, as required for AOCV.
So, e.g. in ``covtype'', for V43 we can see a bias in accuracy with better performance for V43=1 (accuracy=0.973). However, once IFACM is used, this bias is reduced and accuracy for V43=1 becomes closer to the target confidence level 0.95 (accuracy=0.955).
Although improvements in AOCV are observed within all segments, some segments receive greater improvement than others. For example, in ``GPU'', the improved AOCV is much better for KWG than for the segment MWG=64.
\item The code was written in R and was run on a PC with 
a processor
running at 2.1 GHz
with 16GB RAM. The Nelder-Mead optimizer was set to take no more than 100 steps each iteration of the IFACM algorithm. The computation times for training in Algorithm \ref{alg2} on ``covtype'', ``GPU'' and ``KC'' data sets (corresponding to lines 3, 11 and 25 of Table \ref{tab:results}, respectively) were 6125, 225 and 51 seconds respectively. The much longer running time for ``covtype'' was a consequence of a larger training data set size and larger number of iterations, $n_u$.
\end{itemize}

\begin{table}[ht]
\centering
\begin{tabular}{l l l l | l l l l l}
  \hline
Data set & CL & $C$ & & $i=1$ & $i=2$ & $i=3$ & $i=4$ & $i=5$ \\ 
  \hline
covtype & $0.95$ & $0$ & $\gamma_i$ & 0.644 & 0.222 & 0.104 & 0.086 & 0.018 \\
&&& $\delta_i$ & 0.00002 & 0.00015 & 0.00016 & 0.00318 & 0.00089 \\
covtype & $0.95$ & $1$ & $\gamma_i$ & 0.574 & 0.196  \\
&&& $\delta_i$ & 0.00005 & 0.00016 \\
covtype & $0.95$ & $5$ & $\gamma_i$ & 0.448  \\
&&& $\delta_i$ & 0.00008 \\
\hline
GPU & $0.9$ & $0.5$ & $\gamma_i$ & 0.157 & 0.031 & 0.010 \\
&&& $\delta_i$ & 0.00140 & 0.00152 & 0.00063 \\
GPU & $0.9$ & $5$ & $\gamma_i$ & 0.152 & 0.026 \\
&&& $\delta_i$ & 0.00103 & 0.00100\\
\hline
\end{tabular}
\caption{Optimized values of $\gamma_i$ and $\delta_i$ in (\ref{eq:ifacm}) for each iteration $i$ of Algorithm \ref{alg2}, for different experiments.
CL=confidence level $1-\varepsilon$, 
$C$=penalty hyperparameter in (\ref{eq:opt1}),
} 
\label{tab:gamma_delta}
\end{table}

\begin{table}[ht]
\centering
\begin{tabular}{l l l l l l | ll}
  \hline
Data set & CL & $C$ & Variable & Value & \# & Accuracy: & Accuracy: \\ 
&&& (type)&&&Base ICP & IFACM \\
 \hline
covtype & 0.95 & 0 & V1 & $<$median & 50000 & 0.928 & 0.945 \\
& &  & (numeric)  & $\ge$median & 50000 & 0.972 & 0.955 \\
& & & V43 & 0 & 80213 & 0.944 & 0.949 \\
& &  &  (category)& 1 & 19787 & 0.973 & 0.955 \\
\hline
GPU & 0.9 & 0.5 & KWG & 16 & 16212 & 0.933 & 0.896 \\
&&&(category)& 32 & 23788 & 0.874 & 0.902 \\
&&& MWG  & 16 & 3097 & 0.913 & 0.891 \\
&&&(category)& 32 & 7823 & 0.916 & 0.895 \\
&&&& 64 & 12595 & 0.916 & 0.910 \\
&&&& 128 & 16485 & 0.872 & 0.896 \\
\hline
\end{tabular}
\caption{Accuracies within some segments of the test data set using the Base ICP and ICP with IFACM. Each line shows accuracy measured within just the segment of observations given by the value of the variable.
CL=confidence level $1-\varepsilon$,
$C$=penalty hyperparameter in (\ref{eq:opt1}),
\# is the number of observations within the segment.
} 
\label{tab:segments}
\end{table}

\section{Conclusion} \label{conclusion}

The IFACM algorithm is introduced as a mechanism to iteratively update a base CM in order to adjust the conditional probability of a predictive interval being correct so that it is close to the target confidence level across the population and hence achieve AOCV.
The experimental results provide good evidence that IFACM algorithm works well at this task, whilst maintaining marginal validity when the IFACM is used in the ICP.
There is a cost in terms of increased predictive inefficiency, but the impact of that cost differs by data set and for all the data sets used in this study, the cost is not too onerous with the correct choice of hyperparameter $C$.

The results show that deviation from object conditional validity can be high  and hence we recommend that developers using ICP should control for all three objectives: 
\begin{itemize}
\item \emph{Marginal validity}: guaranteed by (\ref{eq:validity}) for any ICP with independent and identically distributed training, calibration and test (or operational) data;
\item \emph{Predictive inefficiency}: needs to be minimized; e.g. minimizing mean prediction interval width (\ref{eq:mean_width}) for regression;
\item \emph{Deviation from conditional validity} (DCV): needs to be minimized; this is the main objective of this study.
\end{itemize}
This article introduces the problem and proposes the novel IFACM algorithm with some initial experimental results that demonstrate that the algorithm is effective. There remain several avenues for further research. 
\begin{enumerate}
\item Further benchmark studies on different data sets, different underlying machine learning algorithms for the base CM and alternative $U$ update functions.
\item Exploring the effect of data set size and model complexity for the overfitting problem identified in the experimental results.
\item Testing alternative probability estimators $p$, such as nonparametric $k$-nearest neighbours, or estimators that could alleviate the overfitting problem, such as those using regularization.
\item The theoretical results given in the Appendix show that the behaviour of the update function is appropriate; however they are not complete and there is room for further theoretical work to help prove that  improved AOCV can be achieved using IFACM.
\item As recommended above, CP should be built taking into account minimizing both predictive inefficiency and DCV. \cite{Colombo2020} and \cite{Bellotti2020} propose methods specifically for minimizing predictive inefficiency, whilst this article focusses just on DCV. It would be interesting to develop a method that integrates both approaches. This is not straightforward since the optimization methods are quite different in each case.
\item This study only considers object conditional validity. It would be interesting to extend to other forms of conditional validity, such as label or example conditional validity.
\end{enumerate}

\bibliographystyle{plainnat}
\bibliography{CPconditional}

\appendix

\section{Theoretical Results}

Some theoretical results for the update function $U^*$ with the logistic regression probability estimator $p$ are given in Theorems 1 and 2 below.
More generally, these results apply to any estimator using a binomial likelihood function (e.g. probit regression).

\newcommand\cprob{\mathbb{P}(C=c \; | \; \bs{x}; \beta)}
\newcommand\codds{\log{\left( \frac{\cprob}{1-\cprob} \right) } }

\begin{proposition} \label{prop1}
Let $S$ be a non-empty finite set and mappings $f: S \to \mathbb{R}$ and $g: S \to \mathbb{R}$.
Let $$\hat\alpha = \arg\max_{\alpha \in S} {f(x)} 
\quad \textrm{and} \quad 
\hat\beta = \arg\max_{\beta \in S} {\left(f(x)+g(x) \right)}.$$
Then, $g(\hat\beta) \ge g(\hat\alpha)$.
\end{proposition}
\begin{proof}
Since $\hat\beta$ maximizes the function, 
$f(\hat\beta)+g(\hat\beta) \ge f(\hat\alpha)+g(\hat\alpha)$
$\ge  f(\hat\beta)+g(\hat\alpha)$
since $\hat\alpha$ maximizes $f$. 
Hence, $g(\hat\beta) \ge g(\hat\alpha)$.
\end{proof}
\begin{definition}
The \emph{binomial log-likelihood} for data set $E$ of observations $(\bs{x}, c) \in E$ with $c \in \{0,1\}$, for some parametric model with coefficients $\beta$, is
$$L(E; \beta) = \sum_{(\bs{x}, y) \in E} {\log \cprob}.$$
\end{definition}
\begin{definition}
The \emph{sample mean of conditional log-odds} of $c \in \{0,1\}$  over data set $E$ of observations $(\bs{x}, c) \in E$ is
$$ \mu(E; \beta) =\frac{1}{|E|} \sum_{(\bs{x}, c) \in E} {\codds}$$
\end{definition}

\begin{proposition} \label{thm:main}
Let $D = \{(\bs{x}_i,c_i) \; | \; i \in \{1,\cdots,n \} \}$ for some $n\ge 1$.\\
Suppose $S \subseteq \{1,\cdots, n\}$. 
Let 
$$D_S = \{(\bs{x}_i,1-c_i) \; | \; i \in S \}, \quad
D_{\bar{S}} = \{(\bs{x}_i,c_i) \in D \; | \; i \not\in S \}, $$
$$\hat\beta_1 = \arg\max_{\beta} {L(D;\beta)} 
\quad \textrm{and} \quad 
\hat\beta_2 = \arg\max_{\beta} {L(D_S \cup D_{\bar{S}} ;\beta)}.$$
Then  
$$\mu(D_S; \hat\beta_2) \ge \mu(D_S; \hat\beta_1).$$
\end{proposition}
\begin{proof}
Let $D_T = \{ (\bs{x}_i, c_i) \in D \; | \; i \in S \}$.
Then since $D_T$ has the same observations as $D_S$ but only differs in outcome,
$$\begin{array}{lll}
L(D_T;\beta) &= \sum_{(\bs{x}, c) \in D_T} {\log\cprob}  \\
&= \sum_{(\bs{x}, c) \in D_S} {\log{\left(\mathbb{P}(C=1-c \; | \; \bs{x}; \beta)  \right)}} \\
&= \sum_{(\bs{x}, c) \in D_S} {\log{\left( 1-\cprob \right)}} & \textrm{since } C\in\{0,1\}.
\end{array}$$
Hence 
$$ L(D_S;\beta)  - L(D_T;\beta) = \sum_{(\bs{x}, y) \in D_S} {\codds} 
= |D_S| \mu(D_S; \beta).$$
Then
$$\begin{array}{lll}
L(D_S \cup D_{\bar{S}} ;\beta) 
&=& L(D_S;\beta)  + L(D_{\bar{S}} ;\beta) \\
&=& L(D_S;\beta)  - L(D_T;\beta) + L(D_T;\beta) + L(D_{\bar{S}} ;\beta) \\
&=& L(D_S;\beta)  - L(D_T;\beta) + L(D;\beta) \\
&=&  L(D;\beta) + |D_S| \mu(D_S; \beta).
\end{array}$$
Then, by Proposition \ref{prop1}, 
$|D_S| \mu(D_S; \hat\beta_2) \ge |D_S| \mu(D_S; \hat\beta_1)$
and we have the result.
\end{proof}

Define the  $\varepsilon$-quantile of the empirical distribution of CM values for the first $k$ examples:
\begin{definition} \label{def:q}
For some base CM $A$ and $\delta\ge0$,
$$q(\gamma)=\min \arg\min_{i \in  \{1,\cdots, k\}} \asgg{\bs{z}_i}$$
such that 
\begin{equation*} 
\sum_{j=1}^k {\mathbb{I}\left[\asgg{\bs{z}_i} 
\ge \asgg{\bs{z}_j} \right] }+1 \ge \varepsilon (k+1).
\end{equation*}
\end{definition}
Notice that the outer $\min$ is needed to deal with the case when two indexes both give the same minimum for the inner $\arg\min$.
\begin{definition}
Let $t(\gamma) = \asgg{\bs{z}_{q(\gamma)}}$.
\end{definition}

\begin{proposition} \label{thm:t}
For all $\gamma\ge 0$,
$t(0)-\gamma \le t(\gamma) \le t(0)+\gamma$.
\end{proposition}
\begin{proof}
\newcommand{\U}[1]{U^*(\bs{z}_#1; \delta, A)}

Let $\gamma_1 < \cdots < \gamma_K$ be all values of $\gamma$, in ascending order, when $q(\gamma)$ changes:
for some $i \neq j$, $\lim_{\eta \to 0^+} q(\gamma+\eta) = i$ and $\lim_{\eta \to 0^-} q(\gamma+\eta) = j$.
Since $A^*_{\gamma}$ is linear on $\gamma$ this must happen only when $\gamma$ is a crossover point; i.e. when 
$\asgg{\bs{z}_i}=\asgg{\bs{z}_j}$.

Prove by induction on ascending values of $\gamma$.
\begin{itemize}
\item Base step, for $\gamma < \gamma_1$:
$q(\gamma)$ remains the same value, i.e. $i=q(\gamma)=q(0)$, so
$t(0)=A(\bs{z}_i)$ and
$$
t(\gamma) = \asgg{\bs{z}_i} 
= A(\bs{z}_i) + \gamma  \U{i}
= t(0) + \gamma \U{i}.
$$
Therefore, $t(0)-\gamma \le t(\gamma) \le t(0)+\gamma$ since $-1 \le \U{i} \le 1$.

\item Inductive step (1), for $\gamma = \gamma_k$ for some $k\ge1$:
let $i=q(\gamma_k)$.\\
$\exists \gamma'<\gamma_k$ with $j=q(\gamma')$ such that
$A^*_{\gamma_k} (\bs{z}_i)=A^*_{\gamma_k} (\bs{z}_j)$.

Then $$\begin{array}{ll}
t(\gamma_k) &= A^*_{\gamma_k} (\bs{z}_j) \\
&= A(\bs{z}_j) + \gamma_k \U{j} \\
&= A^*_{\gamma'}(\bs{z}_j) +  (\gamma_k-\gamma') \U{j} \\
&= t(\gamma') +  (\gamma_k-\gamma') \U{j}
\end{array}$$
$$\Rightarrow  t(\gamma') -  (\gamma_k-\gamma') \le t(\gamma_k) \le t(\gamma') + (\gamma_k-\gamma')$$
since  $-1 \le \U{i} \le 1$. \\
By induction,  $t(0)-\gamma' \le t(\gamma') \le t(0)+\gamma'$
$$\Rightarrow t(0)-\gamma' - (\gamma_k-\gamma') \le t(\gamma_k) \le t(0)+\gamma' + (\gamma_k-\gamma')$$
$$\Rightarrow t(0)-\gamma_k \le t(\gamma_k) \le t(0)+\gamma_k.$$

\item Inductive step (2), for $\gamma>\gamma_k$ for some $k\ge1$ and $\gamma < \gamma_{k+1}$ if  $k<K$: \\
There is no change in $q(\gamma)$ from $\gamma_k$ to $\gamma$, hence $i=q(\gamma)=q(\gamma_k)$.
Therefore,
$$\begin{array}{ll}
t(\gamma)=\asgg{\bs{z}_i}
&=A^*_{\gamma_k}(\bs{z}_i)+(\gamma-\gamma_k) \U{i} \\
&=t(\gamma_k)+(\gamma-\gamma_k) \U{i}.
\end{array}$$
By induction,  $t(0)-\gamma_k \le t(\gamma_k) \le t(0)+\gamma_k$
$$\Rightarrow t(0)-\gamma_k - (\gamma-\gamma_k) \le t(\gamma) \le t(0)+\gamma + (\gamma-\gamma_k)$$
$$\Rightarrow t(0)-\gamma \le t(\gamma) \le t(0)+\gamma.$$
\end{itemize}
\end{proof}

Let the training data set and calibration data sets coincide so that observation $i$ is the same as $i+k$ for $1\le i \le k$ and $l=2k$ for the ICP used in Algorithm \ref{alg1}.
\begin{proposition}
$$\Gamma^{\varepsilon}(\bs{x}) = \left\{y \in  \mathbb{Y} :  \asg{\bs{x}}{y} \ge t(\gamma) \right\}.$$
\end{proposition}
\begin{proof}
Suppose $\asg{\bs{x}}{y} \ge t(\gamma)$, then 
$t(\gamma) \ge \asgg{\bs{z}_j} \Rightarrow \asg{\bs{x}}{y} \ge  \asgg{\bs{z}_j}$
and from Definition \ref{def:q}, it follows that
\begin{equation} \label{eq:prop3proof1}
\sum_{j=1}^k {\mathbb{I}\left[\asg{\bs{x}}{y} 
\ge \asgg{\bs{z}_j} \right] }+1 \ge \varepsilon (k+1).
\end{equation}
Suppose $\asg{\bs{x}}{y} < t(\gamma)$, since $t(\gamma)$ is the smallest value for which Definition  \ref{def:q} holds, then
\begin{equation}  \label{eq:prop3proof2}
\sum_{j=1}^k {\mathbb{I}\left[\asg{\bs{x}}{y} 
\ge \asgg{\bs{z}_j} \right] }+1 < \varepsilon (k+1).
\end{equation}
Since $l=2k$ and all observations $i$ are the same as $i+k$, substituting $A^*_{\gamma}$ for $A$, (\ref{eq:predinterval1}) gives
$$
\Gamma^{\varepsilon}(\bs{x}; A) 
	= \left\{y \in  \mathbb{Y} :  \sum_{j=1}^k {\mathbb{I}(\asg{\bs{x}}{y} \ge \asgg{\bs{z}_j} )} + 1 
  >  \varepsilon (k+1) \right\}
$$
Considering (\ref{eq:prop3proof1}) and  (\ref{eq:prop3proof2}), it immediately follows that
$$\Gamma^{\varepsilon}(\bs{x}) = \left\{y \in  \mathbb{Y} :  \asg{\bs{x}}{y} \ge t(\gamma) \right\}.$$
\end{proof}
The prediction for observation $i$ is correct if and only if
$$y_i \in \Gamma^{\varepsilon}(\bs{x}_i; A^*_{\gamma}) \Leftrightarrow A^*_{\gamma}(\bs{x}_i,y_i) \ge t(\gamma).$$  
Use an indicator variable to express this:
\begin{definition}
Let $c_i(\gamma) = \mathbb{I} [\asgg{\bs{z}_i} \ge t(\gamma) ]$.
\end{definition}
\begin{definition}
Suppose $S$ is an index set of observations and 
let the data set of observations $D_{\gamma} = \{ (\bs{x}_i, c_i(\gamma) ) \; |\; i \in S \}$. Then let 
$$\hat\beta(\gamma) = \arg\max_{\beta} {L(D_{\gamma};\beta)}$$
be the vector of coefficients using logistic regression as probability estimator for the correctness of the prediction intervals.
\end{definition}
We use the notation for the probability estimator in (\ref{eq:probest}) but since we use logistic regression, we supplement it with the coefficients $\beta$ as an additional argument: $p(\bs{z}; A, \beta)$.
For convenience, let $p_i =p(\bs{z}_i; A, \hat\beta(0))$; i.e. the probability estimate from the base ICP using CM $A$.

We now have a framework in place to prove our result.
Our idea is that, for a particular example $\bs{z}_i$, for any feedback-adjusted weight $\gamma>0$ and sensitivity $\delta \ge 0$:
\begin{enumerate}
\item if $c_i(0)=0$ and 
$p_i < 1-\varepsilon$ 
and $c_i(\gamma)=1$, then we expect 
$p(\bs{z}_i; A^*_{\gamma}, \hat\beta(\gamma)) >p_i$;
\item if $c_i(0)=0$ and 
$p_i > (1-\varepsilon) + \delta$, then $c_i(\gamma)=0$;
\item if $c_i(0)=1$ and 
$p_i < (1-\varepsilon) - \delta$, then $c_i(\gamma)=1$;
\item if $c_i(0)=1$ and 
$p_i > 1-\varepsilon$ 
and $c_i(\gamma)=0$, then we expect 
$p(\bs{z}_i; A^*_{\gamma},  \hat\beta(\gamma)) < p_i$.
\end{enumerate}
Line 1 states that if the conditional probability of being correct is biassed below confidence level then if the feedback-adjusted update by $\gamma$ switches it to correct, this should increase the conditional probability of being correct.
Line 4 states a similar objective but, vice versa, for bias above the confidence level.
Both lines 1 and 4 will have the effect of shifting the conditional probability of being correct towards the confidence level.
Line 2 states that if the conditional probability is biassed above confidence level but the outcome is already incorrect, then the feedback-adjusted update for any value $\gamma$ will not change that outcome, since that will not help move the conditional probability towards the confidence level.
Line 3 states a similar requirement, vice versa, for bias below confidence level.
The following Theorem proves lines 2 and 3.
\begin{theorem}
Using update function $U^*$, for $i \in \{1,\cdots,k\}$ and any $\gamma>0$,
\begin{itemize}
\item $c_i(0)=0$ and $ p_i > (1-\varepsilon) + \delta$ implies $c_i(\gamma)=0$.
\item $c_i(0)=1$ and $ p_i < (1-\varepsilon) - \delta$ implies $c_i(\gamma)=1$;
\end{itemize}
\end{theorem}

\begin{proof}
For the first case, $U^*(\bs{z}_i, A, \delta)=-1$ when $p_i > (1-\varepsilon) + \delta$, hence
$$\begin{array}{lll}
c_i(0)=0 & \Rightarrow A(\bs{z}_i) < t(0)  \\
& \Rightarrow A(\bs{z}_i) - \gamma < t(0) - \gamma \le t(\gamma) & \textrm{by Proposition \ref{thm:t}} \\
& \Rightarrow A(\bs{z}_i) + \gamma U^*(\bs{z}_i, A, \delta) <  t(\gamma) &  \\
& \Rightarrow A^*_{\gamma}(\bs{z}_i) <  t(\gamma) &  \\
& \Rightarrow c_i(\gamma)=0 &  
\end{array}$$
For the second case, $U^*(\bs{z}_i, A, \delta)=+1$ when $p_i < (1-\varepsilon) - \delta$, hence
$$\begin{array}{lll}
c_i(0)=1 & \Rightarrow A(\bs{z}_i) > t(0)  \\
& \Rightarrow A(\bs{z}_i) + \gamma > t(0) + \gamma \ge t(\gamma) & \textrm{by Proposition \ref{thm:t}} \\
& \Rightarrow A(\bs{z}_i) + \gamma U^*(\bs{z}_i, A, \delta) >  t(\gamma) &  \\
& \Rightarrow A^*_{\gamma}(\bs{z}_i) >  t(\gamma) &  \\
& \Rightarrow c_i(\gamma)=1 &  
\end{array}$$
\end{proof}

Lines 1 and 4 cannot be proved in the way they are stated, but we can show that, on average, the log-odds of the outcome is not decreasing with $\gamma>0$:
\begin{theorem}
Let $S_{\gamma} = \{(\bs{x}_i,c_i(\gamma)) | i \in \{1,\cdots,k\} \textrm{ and } c_i(0) \ne c_i(\gamma) \}$. 
Then  
$$\mu(S_{\gamma}; \hat\beta(\gamma)) \ge \mu(S_{\gamma}; \hat\beta(0)).$$
\end{theorem}
\begin{proof}
Take $c_i(0)$ as outcome variable $c_i$, $n=k$, and $S=\{i \in \{1,\cdots,k\} | c_i(0) \neq c_i(\gamma)\}$.
Then  since $c_i(\gamma)=1-c_i(0)$ for $i \in S$,
$$D_S=S_{\gamma} \quad \textrm{and} \quad D_{\bar{S}}=\{(\bs{x}_i,c_i(0)) | i \in \{1,\cdots,k\} \textrm{ and } c_i(0) = c_i(\gamma) \}.$$
The result then follows directly from Proposition \ref{thm:main}.
\end{proof}

\end{document}